\documentclass{article}

\usepackage{microtype}
\usepackage{graphicx}
\usepackage{subfigure}
\usepackage{booktabs} 
\usepackage{todonotes}

\usepackage{hyperref}

\usepackage[accepted]{icml2018}

\icmltitlerunning{A Broader View on Bias in Automated Decision-Making}

\begin{document}

\twocolumn[
\icmltitle{A Broader View on Bias in Automated Decision-Making: \\ Reflecting on Epistemology and Dynamics
}



\icmlsetsymbol{equal}{*}

\begin{icmlauthorlist}
\icmlauthor{Roel Dobbe}{eecs}
\icmlauthor{Sarah Dean}{eecs}
\icmlauthor{Thomas Gilbert}{mer}
\icmlauthor{Nitin Kohli}{soi}
\end{icmlauthorlist}

\icmlaffiliation{eecs}{Department of Electrical Engineering and Computer Sciences, University of California Berkeley, USA}
\icmlaffiliation{mer}{Department of Rhetoric, University of California Berkeley, USA}
\icmlaffiliation{soi}{School of Information, University of California Berkeley, USA}

\icmlcorrespondingauthor{Roel Dobbe}{dobbe@berkeley.edu}

\icmlkeywords{Machine Learning, ICML}

\vskip 0.3in
]



\printAffiliationsAndNotice{}  

\begin{abstract}
Machine learning (ML) is increasingly deployed in real world contexts, supplying “actionable insights” and forming the basis of automated decision-making systems. While issues resulting from biases \emph{pre-existing} in training data have been at the center of the fairness debate, these systems are also affected by \emph{technical and emergent biases}, which often arise as context-specific artifacts of implementation. This position paper interprets technical bias as an epistemological problem and emergent bias as a dynamical feedback phenomenon. In order to stimulate debate on how to change machine learning practice to effectively address these issues, we explore this broader view on bias, stress the need to reflect on epistemology, and point to value-sensitive design methodologies to revisit the design and implementation process of automated decision-making systems. 
\end{abstract}

\section{Introduction}

Data-driven decision-making is rapidly being introduced in high-stakes social domains such as medical clinics, criminal justice, and public infrastructure. 
The proliferation of biases in these systems leads to new forms of erroneous decision-making, causing disparate treatment or outcomes across populations~\cite{barocas_big_2016}.
The ML community is working hard to understand and mitigate the unintended
and harmful behavior that may emerge from poor design of real-world automated decision-making systems~\cite{amodei_concrete_2016}.
While many technical tools are being proposed to mitigate these errors, there is insufficient understanding of \emph{how the machine learning design and deployment practice} can safeguard critical human values such as safety or fairness. The AI Now Institute identifies ``a deep need for interdisciplinary, socially aware work that integrates the long history of bias research from the social sciences and humanities into the field of AI research''~\cite{campolo_ai_2017}. 

How can ML practitioners, often lacking consistent language to go beyond technical descriptions and solutions to ``well-defined problems,'' engage with fundamentally human aspects in manner that is \emph{constructive rather than dismissive or reductive}? And how may other disciplines help to enrich the practice?
In this paper, we argue that practitioners and researchers need to take a step back and adopt a broader and more holistic view on bias than currently advocated in many classrooms and professional fora. Our discussion emphasizes the need to reflect on questions of epistemology and underlines the importance of dynamical behavior in data-driven decision-making.
We do not provide full-fledged answers to the problems presented, but point to methodologies in value-sensitive design and self-reflection to contend more effectively with issues of fairness, accountability, and transparency throughout the design and implementation process of automated decision-making systems.

\section{A Broader View On Bias}
Most literature addressing issues of fairness in ML has focused on the ways in which models can inherit \emph{pre-existing biases} from training data. Limiting ourselves to these biases is problematic in two ways.
Firstly, it narrows us to look at how these biases lead to \emph{allocative harm}; a primarily economic view of how systems allocate or withhold an opportunity or resource, such as being granted a loan or held in prison. In her NIPS 2017 keynote, Kate Crawford made the case that at the root of all forms of allocative harm are biases that cause \emph{representational harm}. 
This perspective requires us to move beyond biases in the data set and ``think about the role of ML in harmful representations of human identity,'' and how these biases ``reinforce the subordination of groups along the lines of identity'' and ``affect how groups or individuals are understood socially,'' thereby also contributing to harmful attitudes and cultural beliefs in the longer term~\cite{crawford_keynote:_2017}. 
It is fair to say that representation issues have been largely neglected by the ML community, potentially because they are hard to formalize and track. 

\emph{Responsible representation} requires analyses beyond scrutinizing a training set, including questioning how sensitive attributes might be represented by different features and classes of models and what governance is needed to complement the model.
Additionally, while ML systems are increasingly implemented to provide ``actionable insights'' and guide decisions in the real world, the core methods still fail to effectively address the inherent \emph{dynamic nature} of interactions between the automated decision making process and the environment or individuals that are acted upon. This is particularly true in contexts where observations or human responses (such as clicks and likes) are \emph{fed back} along the way to update the algorithm's parameters, allowing biases to be further reinforced and amplified.

The tendency of ML-based decision-making systems to formalize and reinforce socially sensitive phenomena necessitates a broader taxonomy of biases that includes risks beyond those pre-existing in the data. As argued by Friedman and Nissenbaum in the nascent days of value-sensitive design methodologies, two other sources of bias naturally occur when designing and employing computer systems, namely \emph{technical bias} and \emph{emergent bias}~\cite{friedman_value-sensitive_1996,friedman_bias_1996}. 

While understanding pre-existing bias has lent itself reasonably well to statistical approaches for understanding a given data set, technical and emergent bias require engaging with the domain of application and the ways in which the algorithm is used and integrated. For automated decision-making tools to be responsibly integrated in any context, it is critical that designers (1) assess technical bias by reflecting on their \emph{epistemology} and understanding the values of users and stakeholders, and (2) assess emergent bias by studying the \emph{feedback mechanisms} that create intimate, ever-evolving coupling between algorithms and the environment they act upon.


\section{Technical Bias Is About Epistemology}
Friedman describes a source of technical bias as ``the attempt to make human constructs
amenable to computers - when, for example, we quantify the qualitative, make discrete the continuous, or formalize the nonformal''~\cite{friedman_value-sensitive_1996}. 
This form of bias originates from all the tools used in the process of turning data into a model that can make predictions.
While technical bias is domain-specific, we identify four sources in the machine learning pipeline.

Firstly, both collected and existing data~$X$ are at some point measured and transformed into a computer readable scale. Depending on the objects measured, each variable may have a different scale, such as nominal, ordinal, interval, or ratio.
Consider for example Netflix's decision to let viewers rate movies with ``likes'' instead of a 1-5 star rating. As such, movie ratings moved from an ordinal scale (a number score in which order matters, but the interval between scores does not) to a nominal scale (mutually exclusive labels: you like a movie or you don't).
While the nominal scale might make it easier for viewers to rate movies, it affects how viewers are \emph{represented} and what content gets recommended by the ML system. As such, these choices can produce \emph{measurement bias}, so careful consideration is necessary to understand its effects on system outcomes~\cite{hardt_nips_2017}.

Secondly, based on gathered data~$X$ and available domain knowledge, practitioners \emph{engineer features} and \emph{select model classes}. Features~$\varphi(X)$ can be the available data attributes, transformations thereof based on knowledge and hypotheses, or generated in an automated fashion. Since each feature can be regarded as a model of attributes of the system or population under study, it is relevant to ask how representative it is as a proxy and why it may be predictive of the outcome. Models are used to make predictions based on features. A model class~$f(\cdot;\theta)$, with parameters~$\theta$, should be selected based on the complexity of the phenomenon in question and the amount and quality of the available data.
Is the individual or object that is subject to the decision easily reduced to numbers or equations to begin with? What information in the data is inherently lost by virtue of the mapping $f(\varphi(X);\theta)$ having a limited complexity? 
The process of representation, abstraction and compression can be collectively described as inducing \emph{modeling bias}. 
ML can be seen as a \emph{compression} problem in which complex phenomena are stored as a pattern in a finite-dimensional parameter space. 
From an information theoretic perspective, modeling bias influences the extent to which distortion can be minimized when \emph{reconstructing} a phenomenon from a compressed or sampled version of the original~\cite{cover_elements_2012,dobbe_fully_2017}.

Thirdly, label data $Y$ is used to represent the output of the model. Training labels may be the actual outcome for historical cases, or some discretized or proxy version in cases where the actual outcomes cannot be measured or exactly quantified. Consider for example the use of records of arrest to predict crime rather than the facts of whether the crime was actually committed. How \emph{representative} are such records of real crime across all subpopulations? What core information do they miss for representing the intended classes? And what bias lies hidden in them? We propose to refer to such issues as \emph{label bias}. 

Lastly, given a certain parameterization~$(\varphi(\cdot),f(\cdot,\theta))$ and training data~$(X,Y)$, a model is trained and \emph{tuned to optimize certain objectives}. 
At this stage, various metrics may inform the model builder on where to tweak the model. Do we minimize the number of false positives or false negatives? In recidivism prediction, a false positive may be someone who incorrectly gets sentenced to prison, whereas a false negative poses a threat to safety by failing to recognize a high-threat individual. There are inherent trade-offs between prioritizing for equal prediction accuracy across groups versus for an equal likelihood of false positives and negatives across groups~\cite{chouldechova_fair_2016,kleinberg_inherent_2017}.
Technical definitions of fairness are motivated by different metrics, illuminating the inherent ambiguity and context-dependence of such issues.
For a given context, what is the right balance? And who gets to decide? We coin the effects of these trade-offs \emph{optimization bias}.

The many questions posed above illuminate the range of places in the machine learning design process where issues of \emph{epistemology} arise: they require \emph{justification} and often \emph{value judgment}.
Our theory of knowledge and the way we formalize and solve problems determines how we represent and understand sensitive phenomena.
How do we represent phenomena in ways that are deemed correct? What evidence is needed in order to justify an action or decision? What are legitimate classes or outcomes of a model? And how do we deal with inherent trade-offs of fairness? These challenges are deeply context-specific, often ethical, and challenge us to understand our epistemology and that of the domain we are working in. 

The detrimental effects of overlooking these questions in practice are obvious in high-stakes domains, such as predictive policing and sentencing, where the decision to treat crime as a prediction problem reduces the perceived autonomy of individuals, fated to either commit a crime or act within the law. 
Barabas et al. argue that rather than prediction, ``machine learning models should be used to surface covariates that are fed into a \emph{causal model} for understanding the social, structural and psychological drivers of crime''~\cite{barabas_interventions_2018}. This is a strong message with many challenges, but it points in the right direction: in these contexts,  machine learning models should \emph{facilitate rather than replace} the critical eye of the human expert. It forces practitioners and researchers to be humble and reflect on how \emph{our own skills and tools may benefit or hurt an existing decision-making process}.

\section{Emergent Bias Is About Dynamics}
\begin{figure}
    \centering
    \includegraphics[width=0.45\textwidth]{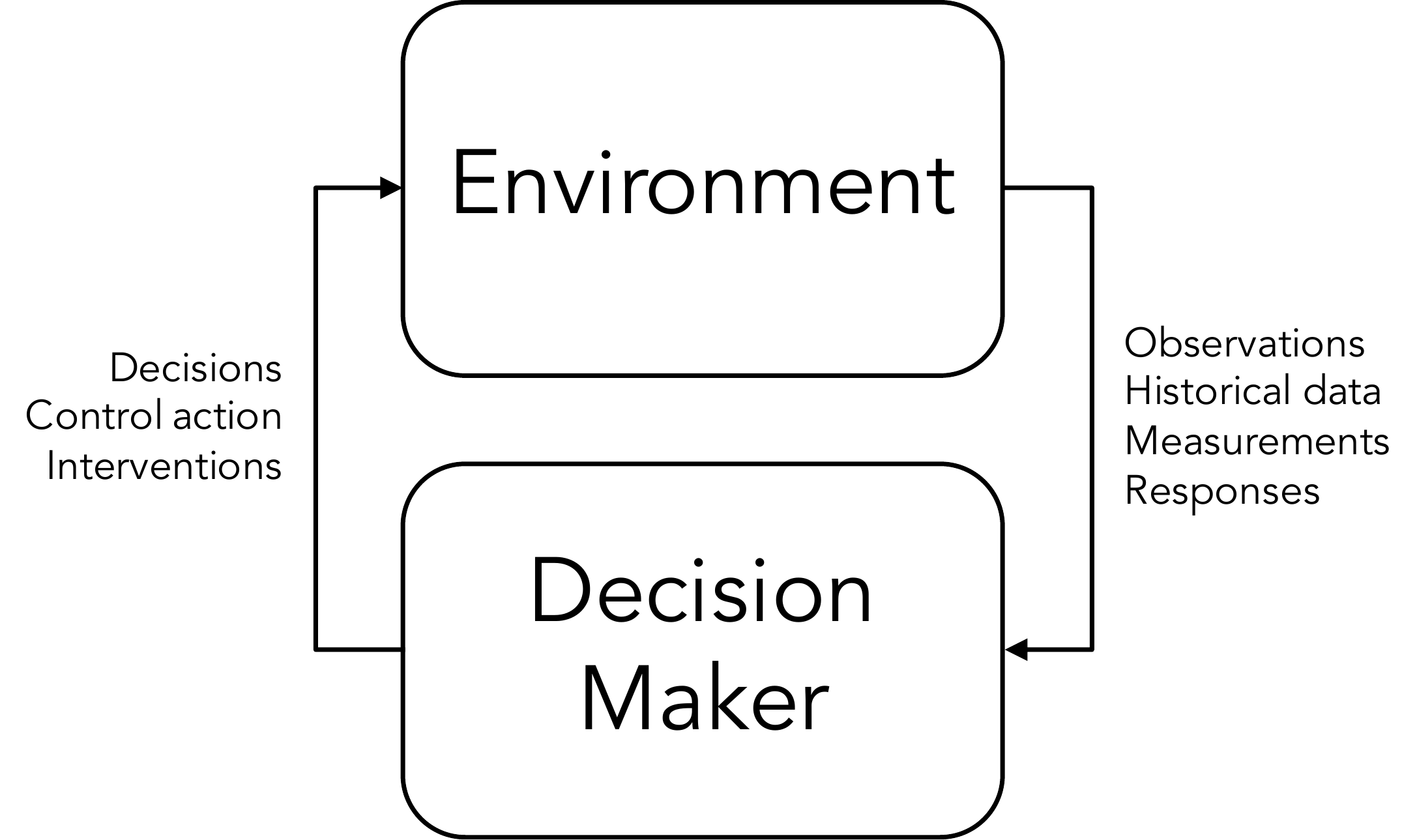}
    \caption{A Simple Feedback Model}
    \label{fig:my_label}
\end{figure}

Complementing pre-existing and technical biases, ``emergent bias arises only in a context of use by real users [...] as a result of a change in societal knowledge, user population, or cultural values.''~\cite{friedman_value-sensitive_1996}.
Recently, convincing examples of emergent bias have surfaced in contexts where ML is used to automate or mediate human decisions. 
In predictive policing, where discovered crime data (e.g., arrest records)
are used to predict the location of new crimes and determine police deployment, runaway feedback loops can cause increasing surveillance of particular neighborhoods regardless of the true crime rate~\cite{ensign_runaway_2018}, leading to over-policing of ``high-risk'' individuals~\cite{stroud_chicagos_2016}.
In optimizing for attention, recommendation systems may have a tendency to turn towards the extreme and radical~\cite{tufekci_opinion_2018}.
When machine learning systems are unleashed in feedback with society, they may be more accurately described as \emph{reinforcement learning} systems, performing \emph{feedback control}~\cite{recht_ethics_2018}.
Therefore, a decision-making system has its own \emph{dynamics}, which can be modified by feedback, potentially causing bias to accrue over time. To conceptualize these ideas at a high level, we adopt the system formulation depicted in Figure~\ref{fig:my_label}.

The machine learning system acts on the environment through decisions, control actions, or interventions. From the environment, the decision maker considers observations, historical data, measurements and responses, conceivably updating its model in order to steer the environment in a beneficial direction.  For example, in the case of predictive policing, `the environment' describes a city and its citizens, and `the decision maker' is the police department, which determines where to send police patrols or invest in social interventions.

The dynamical perspective offered by the conception of a feedback model allows for a focus on interactions, which can add clarity to debates over key issues like fairness and algorithmic accountability. Situations with completely different fairness interpretations may have identical \emph{static} observational metrics (properties of the joint distribution of input, model and output), and thus a causal or dynamic model is necessary to distinguish them~\cite{hardt_equality_2016}. 
On the other hand, a one-step feedback model, incorporating temporal indicators of well-being for individuals affected by decisions,
offers a way of comparing competing definitions of fairness~\cite{liu_delayed_2018}.
Similarly, calls for ``interpretability'' and proposed solutions often omit key operative words -- Interpretable to whom? And for what purpose?~\cite{kohli_translation_2018}. The dynamic viewpoint adds clarity to these questions by focusing on causes and effects of decision making systems, and situating interpretability in context.

Beyond providing a more realistic and workable frame of thinking about bias and related issues, the feedback system perspective may also allow inspiration to be drawn from areas of \emph{Systems Theory} that have traditionally studied feedback and dynamics. For instance, the field of \emph{System Identification} uses statistical methods to build mathematical models of dynamical systems from measured data, often to be employed to control \emph{dynamical systems} with strict safety requirements, such as airplanes or electric power systems~\cite{ljung_system_1998,astrom_feedback_2010}. 
Inspiration may be drawn from the rich literature on \emph{closed-loop identification}, which considers the identification of models with data gathered \emph{during} operation, while the same model is also used to safeguard a system~\cite{van_den_hof_closed-loop_1998}. 
That said, modeling socio-technical systems is more challenging than engineered systems. The complexity of modeled phenomena, the role of unmodeled phenomena such as external economic factors, and slower temporal dynamics all pose barriers to directly applying existing engineering principles.

\section{Our Positionality Shapes Our Epistemology}
As ML practitioners and researchers, we are wired to analyze challenges in ways that \emph{abstract, formalize and reduce complexity}. 
It is natural for us to think rigorously about technical roots of biases in the systems we design, and propose and techno-fixes to prevent negative impact from their proliferation. However, it is of crucial importance to acknowledge that the methods and approaches we use to reduce, formalize, and gather feedback from experiments are \emph{themselves} inherent sources of bias. 
Epistemologies differ tremendously from application to application and ultimately shape the way a decision-maker justifies decisions and affects individuals. 
Technology \emph{intimately touches and embodies values} deemed critical in employing the intended decision-making system. 
As such, we need to go beyond our formal tools and analyses to engage with others and reflect on our epistemology. In doing so, we aim to determine \emph{responsible ways} in which technology can help put values into practice, and understand its fundamental limitations.

With a plethora of issues surfacing, it is easy to either consider banning ML altogether, or otherwise dismiss requests to fundamentally revisit its role in enabling data-driven decision-making in sensitive environments.
Instead, we propose three principles to nourish debate on the middle ground:

\textbf{1}: \emph{Do fairness forensics}~\cite{crawford_keynote:_2017}. Keep track of biases in an open and transparent way and engage in constructive dialogue with domain experts, to understand proven ways of formalizing complex phenomena and to breed awareness about how bias works and when/where users should be cautious.

\textbf{2}: \emph{Acknowledge that your positionality shapes your epistemology}~\cite{takacs_how_2003}. Our personal backgrounds, the training we received, the people we represent or interact with all have an impact on how we look at and formalize problems. As ML practitioners, we should set aside time and energy for critical self-reflection, to identify our own biases and blind spots, to harbor communication with the groups affected by the systems we design, and to understand where we should enrich our epistemology with other viewpoints.

\textbf{3}: \emph{Perform value-sensitive design}. Determine what values are relevant in building a decision-making system and how they might be embodied or challenged in the design and implementation by engaging with users and other affected stakeholders~\cite{van_den_hoven_ict_2007,friedman_value_2013}.

As Takacs describes it, the benefits of self-reflection go well beyond arriving at the ``best solution'' to a complex problem~\cite{takacs_how_2003,takacs_positionality_2002}. ``This means learning to listen with open minds and hearts, learning to respect different ways of knowing the world borne of different identities and experiences, and learning to examine and re-examine one's own worldviews. [...] When we constantly engage to understand how our positionality biases our epistemology, we greet the world with respect, interact with others to explore and cherish their differences, and live life with a fuller sense of self as part of a web of community.''

As machine learning systems rapidly change our information gathering and shape our decisions and worldviews in ways we cannot fully anticipate, self-reflection and awareness of our epistemology becomes ever more important for machine learning practitioners and researchers to ensure that automated decision-making systems contribute in beneficial and sustainable ways.

\paragraph{Acknowledgements}

We thank Moritz Hardt and Ben Recht for helpful comments and suggestions.
This work is funded by a Tech for Social Good Grant from CITRIS and the Banatao Institute at UC Berkeley.

\bibliography{references}

\begin{thebibliography}{26}
\providecommand{\natexlab}[1]{#1}
\providecommand{\url}[1]{\texttt{#1}}
\expandafter\ifx\csname urlstyle\endcsname\relax
  \providecommand{\doi}[1]{doi: #1}\else
  \providecommand{\doi}{doi: \begingroup \urlstyle{rm}\Url}\fi

\bibitem[Amodei et~al.(2016)Amodei, Olah, Steinhardt, Christiano, Schulman, and
  Man{\'e}]{amodei_concrete_2016}
Amodei, Dario, Olah, Chris, Steinhardt, Jacob, Christiano, Paul, Schulman,
  John, and Man{\'e}, Dan.
\newblock Concrete {Problems} in {AI} {Safety}.
\newblock \emph{arXiv:1606.06565 [cs]}, June 2016.
\newblock URL \url{http://arxiv.org/abs/1606.06565}.
\newblock arXiv: 1606.06565.

\bibitem[{\AA}str{\"o}m \& Murray(2010){\AA}str{\"o}m and
  Murray]{astrom_feedback_2010}
{\AA}str{\"o}m, Karl~Johan and Murray, Richard~M.
\newblock \emph{Feedback systems: an introduction for scientists and
  engineers}.
\newblock Princeton university press, 2010.

\bibitem[Barabas et~al.(2018)Barabas, Dinakar, Virza, and
  Zittrain]{barabas_interventions_2018}
Barabas, Chelsea, Dinakar, Karthik, Virza, Joichi Ito~Madars, and Zittrain,
  Jonathan.
\newblock Interventions over {Predictions}: {Reframing} the {Ethical} {Debate}
  for {Actuarial} {Risk} {Assessment}.
\newblock In \emph{1st Conference on Fairness, Accountability, and
  Transparancy}, New York, NY, USA, February 2018.
\newblock URL \url{http://arxiv.org/abs/1712.08238}.
\newblock arXiv: 1712.08238.

\bibitem[Barocas \& Selbst(2016)Barocas and Selbst]{barocas_big_2016}
Barocas, Solon and Selbst, Andrew~D.
\newblock Big data's disparate impact.
\newblock \emph{Cal. L. Rev.}, 104:\penalty0 671, 2016.

\bibitem[Campolo et~al.(2017)Campolo, Sanfilippo, Whittaker, and
  Crawford]{campolo_ai_2017}
Campolo, Alex, Sanfilippo, Madelyn, Whittaker, Meredith, and Crawford, Kate.
\newblock {AI} {Now} 2017 {Report}.
\newblock Technical report, AI Now Institute, New York, NY, USA, 2017.

\bibitem[Chouldechova(2016)]{chouldechova_fair_2016}
Chouldechova, Alexandra.
\newblock Fair prediction with disparate impact: {A} study of bias in
  recidivism prediction instruments.
\newblock \emph{arXiv:1610.07524 [cs, stat]}, October 2016.
\newblock URL \url{http://arxiv.org/abs/1610.07524}.
\newblock arXiv: 1610.07524.

\bibitem[Cover \& Thomas(2012)Cover and Thomas]{cover_elements_2012}
Cover, Thomas~M. and Thomas, Joy~A.
\newblock \emph{Elements of information theory}.
\newblock John Wiley \& Sons, 2012.

\bibitem[Crawford(2017)]{crawford_keynote:_2017}
Crawford, Kate.
\newblock Keynote: {The} {Trouble} with {Bias}, December 2017.
\newblock URL \url{https://www.youtube.com/watch?v=fMym_BKWQzk}.

\bibitem[Dobbe et~al.(2017)Dobbe, Fridovich-Keil, and Tomlin]{dobbe_fully_2017}
Dobbe, Roel, Fridovich-Keil, David, and Tomlin, Claire.
\newblock Fully {Decentralized} {Policies} for {Multi}-{Agent} {Systems}: {An}
  {Information} {Theoretic} {Approach}.
\newblock In \emph{Conference on Neural Information Processing Systems}, Long
  Beach, CA, USA, December 2017.
\newblock URL \url{https://arxiv.org/abs/1707.06334}.

\bibitem[Ensign et~al.(2018)Ensign, Friedler, Neville, Scheidegger, and
  Venkatasubramanian]{ensign_runaway_2018}
Ensign, Danielle, Friedler, Sorelle~A., Neville, Scott, Scheidegger, Carlos,
  and Venkatasubramanian, Suresh.
\newblock Runaway feedback loops in predictive policing.
\newblock In \emph{Conference on {Fairness}, {Accountability} and
  {Transparency}}, New York, NY, USA, February 2018.

\bibitem[Friedman(1996)]{friedman_value-sensitive_1996}
Friedman, Batya.
\newblock Value-sensitive {Design}.
\newblock \emph{interactions}, 3\penalty0 (6):\penalty0 16--23, December 1996.
\newblock ISSN 1072-5520.
\newblock \doi{10.1145/242485.242493}.
\newblock URL \url{http://doi.acm.org/10.1145/242485.242493}.

\bibitem[Friedman \& Nissenbaum(1996)Friedman and
  Nissenbaum]{friedman_bias_1996}
Friedman, Batya and Nissenbaum, Helen.
\newblock Bias in {Computer} {Systems}.
\newblock \emph{ACM Trans. Inf. Syst.}, 14\penalty0 (3):\penalty0 330--347,
  July 1996.
\newblock ISSN 1046-8188.
\newblock \doi{10.1145/230538.230561}.
\newblock URL \url{http://doi.acm.org/10.1145/230538.230561}.

\bibitem[Friedman et~al.(2013)Friedman, Kahn~Jr, Borning, and
  Huldtgren]{friedman_value_2013}
Friedman, Batya, Kahn~Jr, Peter~H., Borning, Alan, and Huldtgren, Alina.
\newblock Value sensitive design and information systems.
\newblock In \emph{Early engagement and new technologies: {Opening} up the
  laboratory}, pp.\  55--95. Springer, 2013.

\bibitem[Hardt \& Barocas(2017)Hardt and Barocas]{hardt_nips_2017}
Hardt, Moritz and Barocas, Solon.
\newblock {NIPS} 2017 - {Tutorial} on {Fairness} in {Machine} {Learning},
  December 2017.
\newblock URL \url{http://mrtz.org/nips17/}.

\bibitem[Hardt et~al.(2016)Hardt, Price, and Srebro]{hardt_equality_2016}
Hardt, Moritz, Price, Eric, and Srebro, Nati.
\newblock Equality of opportunity in supervised learning.
\newblock In \emph{Advances in {Neural} {Information} {Processing} {Systems}},
  pp.\  3315--3323, 2016.

\bibitem[Kleinberg et~al.(2017)Kleinberg, Mullainathan, and
  Raghavan]{kleinberg_inherent_2017}
Kleinberg, Jon, Mullainathan, Sendhil, and Raghavan, Manish.
\newblock Inherent {Trade}-{Offs} in the {Fair} {Determination} of {Risk}
  {Scores}.
\newblock In \emph{8th Innovations in Theoretical Computer Science Conference
  (ITCS 2017)}, volume~67, pp.\  43:1--43:23, 2017.
\newblock arXiv: 1609.05807.

\bibitem[Kohli et~al.(2018)Kohli, Barreto, and Kroll]{kohli_translation_2018}
Kohli, Nitin, Barreto, Renata, and Kroll, Joshua~A.
\newblock Translation {Tutorial}: {A} {Shared} {Lexicon} for {Research} and
  {Practice} in {Human}-{Centered} {Software} {Systems}.
\newblock In \emph{1st Conference on Fairness, Accountability, and
  Transparancy}, pp.\ ~7, New York, NY, USA, February 2018.

\bibitem[Liu et~al.(2018)Liu, Dean, Rolf, Simchowitz, and
  Hardt]{liu_delayed_2018}
Liu, Lydia~T., Dean, Sarah, Rolf, Esther, Simchowitz, Max, and Hardt, Moritz.
\newblock Delayed {Impact} of {Fair} {Machine} {Learning}.
\newblock \emph{arXiv:1803.04383 [cs, stat]}, March 2018.
\newblock URL \url{http://arxiv.org/abs/1803.04383}.
\newblock arXiv: 1803.04383.

\bibitem[Ljung(1998)]{ljung_system_1998}
Ljung, Lennart.
\newblock \emph{System identification}.
\newblock Springer, 1998.
\newblock URL
  \url{http://link.springer.com/chapter/10.1007/978-1-4612-1768-8_11}.

\bibitem[Recht(2018)]{recht_ethics_2018}
Recht, Benjamin.
\newblock The {Ethics} of {Reward} {Shaping}, April 2018.
\newblock URL
  \url{http://benjamin-recht.github.io/2018/04/16/ethical-rewards/}.

\bibitem[Stroud(2016)]{stroud_chicagos_2016}
Stroud, Matt.
\newblock Chicago's predictive policing tool just failed a major test, August
  2016.
\newblock URL
  \url{https://www.theverge.com/2016/8/19/12552384/chicago-heat-list-tool-failed-rand-test}.

\bibitem[Takacs(2002)]{takacs_positionality_2002}
Takacs, David.
\newblock Positionality, epistemology, and social justice in the classroom.
\newblock \emph{Social Justice}, 29\penalty0 (4 (90):\penalty0 168--181, 2002.

\bibitem[Takacs(2003)]{takacs_how_2003}
Takacs, David.
\newblock How does your positionality bias your epistemology?
\newblock \emph{Thought \& Action}, 27, 2003.

\bibitem[Tufekci(2018)]{tufekci_opinion_2018}
Tufekci, Zeynep.
\newblock Opinion {\textbar} {YouTube}, the {Great} {Radicalizer}.
\newblock \emph{The New York Times}, March 2018.
\newblock ISSN 0362-4331.
\newblock URL
  \url{https://www.nytimes.com/2018/03/10/opinion/sunday/youtube-politics-radical.html}.

\bibitem[Van~den Hof(1998)]{van_den_hof_closed-loop_1998}
Van~den Hof, Paul.
\newblock Closed-loop issues in system identification.
\newblock \emph{Annual reviews in control}, 22:\penalty0 173--186, 1998.

\bibitem[Van~den Hoven(2007)]{van_den_hoven_ict_2007}
Van~den Hoven, Jeroen.
\newblock {ICT} and value sensitive design.
\newblock \emph{The information society: Innovation, legitimacy, ethics and
  democracy in honor of Professor Jacques Berleur SJ}, pp.\  67--72, 2007.

\end{thebibliography}
\bibliographystyle{icml2018}
\end{document}